\documentclass[11pt]{article}

\usepackage[preprint]{acl}
\usepackage{booktabs, makecell}
\usepackage[most]{tcolorbox}
\usepackage{url}
\usepackage{enumitem}
\usepackage{amsmath, amssymb}
\usepackage{array}
\usepackage{booktabs}
\usepackage{multirow}
\usepackage{times}
\usepackage{latexsym}
\usepackage{enumitem}

\usepackage[T1]{fontenc}

\usepackage[utf8]{inputenc}

\usepackage{microtype}

\usepackage{inconsolata}

\usepackage{graphicx}
\usepackage[table]{xcolor}  
\usepackage{colortbl}       

\definecolor{em}{gray}{0.9}

\usepackage{colortbl}
\definecolor{bestbg}{RGB}
{255,235,235}   
\definecolor{thirdbg}{RGB}{210, 240, 210}
\definecolor{secondbg}{RGB}{235,245,255} 
\definecolor{qblue}{RGB}{230,245,255}
\usepackage{tabularx,booktabs}
\newcolumntype{L}[1]{>{\raggedright\arraybackslash}p{#1}}
\newcolumntype{Y}{>{\raggedright\arraybackslash}X}
\usepackage{xcolor}
\usepackage{soul}
\definecolor{lightpink}{HTML}{F7DADA}

\title{NeuronScope: A Multi-Agent Framework for Explaining Polysemantic Neurons in Language Models}

\author{
  Weiqi Liu$^{1}$ \quad
  Yongliang Miao$^{2}$ \quad
  Haiyan Zhao$^{3}$ \quad
  Yanguang Liu$^{3}$ \quad
  Mengnan Du$^{4}$\textsuperscript{†}\\[4pt]
  $^{1}$Wuhan University \quad
  $^{2}$Hong Kong Baptist University \quad \\
  $^{3}$New Jersey Institute of Technology \quad
  $^{4}$The Chinese University of Hong Kong, Shenzhen \\[4pt]
  \small\texttt{michaelliuwq003@gmail.com}, \;
  \small\texttt{r130026108@gmail.com} \\
  \small\texttt{\{hz54,yanguang.liu\}@njit.edu}, \;
  \small\texttt{mengnandu@cuhk.edu.cn} \\
  \textsuperscript{†}Corresponding author
}

\begin{document}
\maketitle
\begin{abstract}
Neuron-level interpretation in large language models (LLMs) is fundamentally challenged by widespread polysemanticity, where individual neurons respond to multiple distinct semantic concepts. Existing single-pass interpretation methods struggle to faithfully capture such multi-concept behavior. In this work, we propose \textbf{NeuronScope}, a multi-agent framework that reformulates neuron interpretation as an iterative, activation-guided process. NeuronScope explicitly deconstructs neuron activations into atomic semantic components, clusters them into distinct semantic modes, and iteratively refines each explanation using neuron activation feedback. Experiments demonstrate that NeuronScope uncovers hidden polysemanticity and produces explanations with significantly higher activation correlation compared to single-pass baselines.
\end{abstract}


\section{Introduction}
Large language models (LLMs) have achieved remarkable progress across diverse tasks, including natural language understanding, generation, and reasoning. Despite their impressive capabilities, LLMs remain largely opaque systems whose internal mechanisms are poorly understood~\cite{zhao2024explainability}. To address this opacity, the research community has increasingly focused on decoding LLM representations to understand how these models process and store knowledge. A primary avenue of investigation involves interpreting individual processing units, such as neurons in Multi-Layer Perceptrons (MLPs) or features extracted via Sparse Autoencoders (SAEs), in natural language~\cite{bills2023language,han2025sageagenticexplainerframework}. 

\begin{figure*}[tb!]
\centering{
\includegraphics[width=0.99\textwidth]{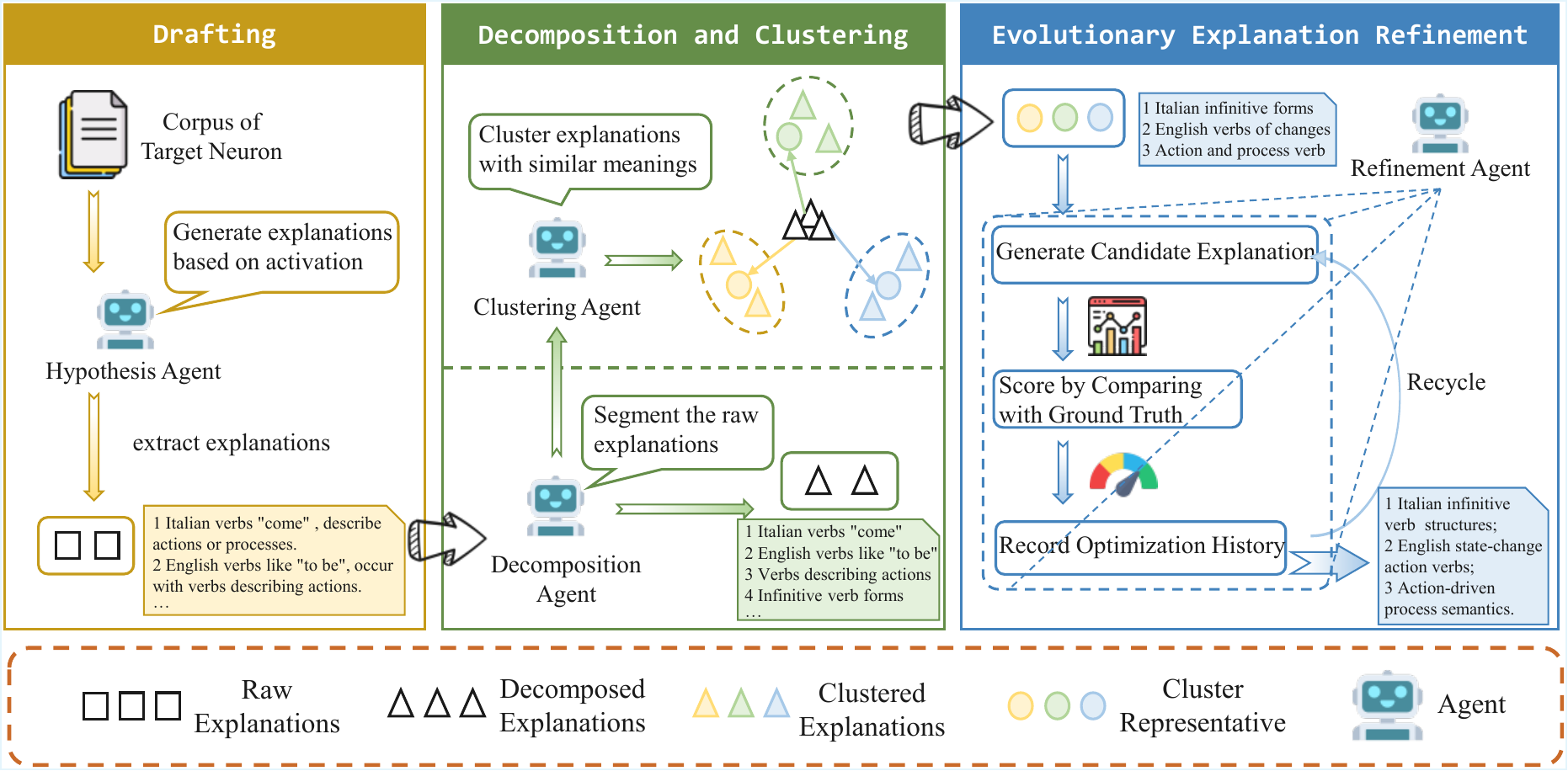}}

\caption{The architecture of NeuronScope. (1) Initial Explanation, where a Hypothesis Agent generates broad raw explanations from high-activation corpus exemplars, prioritizing recall; (2) Decomposition and Clustering, where a Decomposition Agent breaks composite hypotheses into atomic components which are then grouped into distinct semantic modes by a Clustering Agent ; and (3) Evolutionary Explanation Refinement, where a Refinement Agent treats explanation generation as an optimization problem, iteratively updating descriptions based on activation feedback to maximize precision.}
\label{fig:main_figure}
\end{figure*}

Although automatic generation of natural language descriptions for neurons has been demonstrated, two fundamental limitations persist. First, existing studies focus on enforcing a single description, yielding vague generalizations that fail to capture specific activation patterns.
In practice, many neurons are polysemantic, activating upon multiple, distinct, and often unrelated semantic or structural patterns~\cite{elhage2022toymodelssuperposition}. 
Second, existing generation methods rely on one-shot generation or simple best-of-k sampling, lacking a feedback loop to iteratively refine the explanation against ground-truth activation data, thereby limiting the predictive power of the resulting interpretations.

To address these challenges, we propose \textbf{NeuronScope}, a multi-agent framework that reframes feature interpretation from a one-shot generation problem into a structured, iterative refinement process. 
Rather than forcing a single summary of complex behaviors, NeuronScope orchestrates role-specialized agents for hypothesis proposing, semantic decomposition, and refinement, to systematically observe raw activation patterns and disentangle mixed semantics into distinct atomic components. 
Building on these components, NeuronScope further leverages activation-guided feedback to iteratively revise explanations, yielding more faithful semantic disentanglement and precise, multi-faceted interpretations. 
We perform experiments across three open-source LLMs, demonstrating that NeuronScope successfully disentangles hidden polysemanticity and produces explanations with higher activation correlation scores compared to existing single-pass baselines.

\section{Methodology}


\subsection{Problem Formulation}

Individual neurons serve as basic units of mechanistic interpretability in LLMs. 
As demonstrated, many neurons exhibit polysemantic behavior, activating on multiple semantically distinct patterns.

Consider a neuron $n$ that maps an input text $x$ to a scalar activation $a_n(x) \in \mathbb{R}$. We characterize a neuron's behavior using its high-activation dataset $\mathcal{D}$, defined as the set of text segments that induce large values of $a_n(x)$. 
For polysemantic neurons, $\mathcal{D}$ can be viewed as a mixture of $K$ latent semantic components, $\mathcal{D} = \{\mathcal{D}_k\}_{k=1}^{K}$, where $K\geq 1$ and each subset $\mathcal{D}_k$ corresponds to text segments that activate the neuron due to a distinct concept.


Our goal is to generate precise natural language explanations for each individual subset. 
Let $\mathcal{L}$ denote the space of candidate textual descriptions. 
We seek a set of explanations $\{E_1, \ldots, E_K\}$, where each explanation $E_k \in \mathcal{L}$ is associated with one component $\mathcal{D}_k$. For each component $k$, the target explanation is defined as
\begin{equation}
E_k^* = \arg\max_{E \in \mathcal{L}} \operatorname{Score}(E, \mathcal{D}_k),
\end{equation}
where $\operatorname{Score}(E, \mathcal{D}_k)$ quantifies how faithfully an explanation $E$ predicts the neuron's activation on held-out data. 

\subsection{Explanation Drafting}

Before generating explanations, we first extract text segments from the corpus that elicit the highest activations for the neuron, forming an empirical reference set of its behavior. Then, the Hypothesis Agent is applied to these examples to generate a comprehensive Raw Explanation, denoted as $H_{\text{raw}}$. Rather than prematurely collapsing behavior into one dominant concept, the proposer deliberately retains all plausible cues to support downstream decomposition and grouping.


\subsection{Semantic Decomposition and Clustering}

To generate comprehensive explanations, NeuronScope employs a two-stage, agent-driven factorization, where a Decomposition Agent extracts atomic conditions from $H_{\text{raw}}$, and a Clustering Agent aggregates them into semantically coherent groups.

\noindent
\textbf{Atomic Decomposition.} The Decomposition Agent processes the composite hypothesis $H_{\text{raw}}$ to produce a set of discrete, atomic components $\{c_1, c_2, \dots, c_m\}$. 
Each atomic component represents a single, falsifiable condition.

\noindent
\textbf{Multi-Semantic Clustering.} To eliminate redundancy, the Clustering Agent projects these atomic components into an embedding space using an embedding model. The Clustering Agent applies a clustering algorithm to aggregate components into $N$ distinct Semantic Clusters $\{S_1, S_2, ..., S_N\}$. Each cluster represents a unique semantic facet of the neuron. For instance, a single neuron may be resolved into distinct clusters corresponding to contextual causation and prepositional usage, which are subsequently treated as separate interpretation targets.

\begin{table*}[tp]
  \centering
  \caption{Performance of NeuronScope and baseline in terms of neuron-level Score and Number metrics.}
  \label{tab:explanation_quality}
  \resizebox{0.8\linewidth}{!}{%
  \begin{tabular}{@{}lccccccccc@{}}
    \toprule
    \textbf{Method} & \multicolumn{3}{c}{\textbf{Llama-3.1-8b}} & \multicolumn{3}{c}{\textbf{Gemma-2-2b}} & \multicolumn{3}{c}{\textbf{Qwen3-4b}} \\
    \cmidrule(lr){2-4} \cmidrule(lr){5-7} \cmidrule(lr){8-10}
    & Layer & Score$\uparrow$ & Number & Layer & Score$\uparrow$ & Number & Layer & Score$\uparrow$ & Number \\
    \midrule
    NeuronScope (w/o Refinement) & 5 & 0.453 & 5.05 & 5 & 0.600 & 2.48 & 5 & 0.489 & 2.29 \\
    \rowcolor{qblue} \textbf{NeuronScope} & 5 & \textbf{0.508} & 5.05 & 5 & \textbf{0.663} & 2.48 & 5 & \textbf{0.589} & 2.29 \\
    \midrule
    NeuronScope (w/o Refinement) & 10 & 0.334 & 5.53 & 9 & 0.539 & 2.49 & 9 & 0.469 & 2.37 \\
    \rowcolor{qblue} \textbf{NeuronScope} & 10 & \textbf{0.390} & 5.53 & 9 & \textbf{0.620} & 2.49 & 9 & \textbf{0.567} & 2.37 \\
    \midrule
    NeuronScope (w/o Refinement) & 20 & 0.464 & 4.70 & 18 & 0.517 & 2.22 & 18 & 0.461 & 2.26 \\
    \rowcolor{qblue} \textbf{NeuronScope} & 20 & \textbf{0.516} & 4.70 & 18 & \textbf{0.619} & 2.22 & 18 & \textbf{0.562} & 2.26 \\
    \midrule
    NeuronScope (w/o Refinement) & 31 & 0.364 & 4.48 & 25 & 0.452 & 2.32 & 25 & 0.475 & 2.42 \\
    \rowcolor{qblue} \textbf{NeuronScope} & 31 & \textbf{0.411} & 4.48 & 25 & \textbf{0.550} & 2.32 & 25 & \textbf{0.574} & 2.42 \\
    \bottomrule
  \end{tabular}%
  }
\end{table*}

\subsection{Evolutionary Explanation Refinement}

The final stage utilizes the Refinement Agent to optimize explanations in an iterative manner.
For each semantic cluster $S_i$, the Refinement Agent selects a representative explanation and iteratively optimizes it to maximize $\operatorname{Score}(E, \mathcal{D}_k)$, which measures the correlation between activations of explanation $E$ and the neuron’s ground-truth activations. The process proceeds as follows:
\begin{itemize}
[leftmargin=10pt, topsep=-2pt, itemsep=1pt, partopsep=1pt, parsep=1pt] 
    \item \textbf{Scoring:} The current explanation $E^{(t)}$ is evaluated on held-out text sequences to obtain its activation-prediction score.
    \item \textbf{Refinement:} The agent reviews the optimization trajectory upon the history of explanations and scores $\{ (E^{(0)}, \text{Score}^{(0)}), \ldots, (E^{(t)}, \text{Score}^{(t)}) \}$. Acting as an optimizer, the agent proposes a new candidate explanation $E^{(t+1)}$ 
    by refining linguistic specificity to better capture observed activation patterns.
\end{itemize}

This loop continues for a fixed number of iterations or until convergence. By effectively performing gradient ascent in natural language space, this procedure yields explanations that are both descriptively precise and maximally predictive of the neuron's internal behavior.

\begin{figure*}[htbp]
    \centering
    \includegraphics[width=0.85\linewidth]{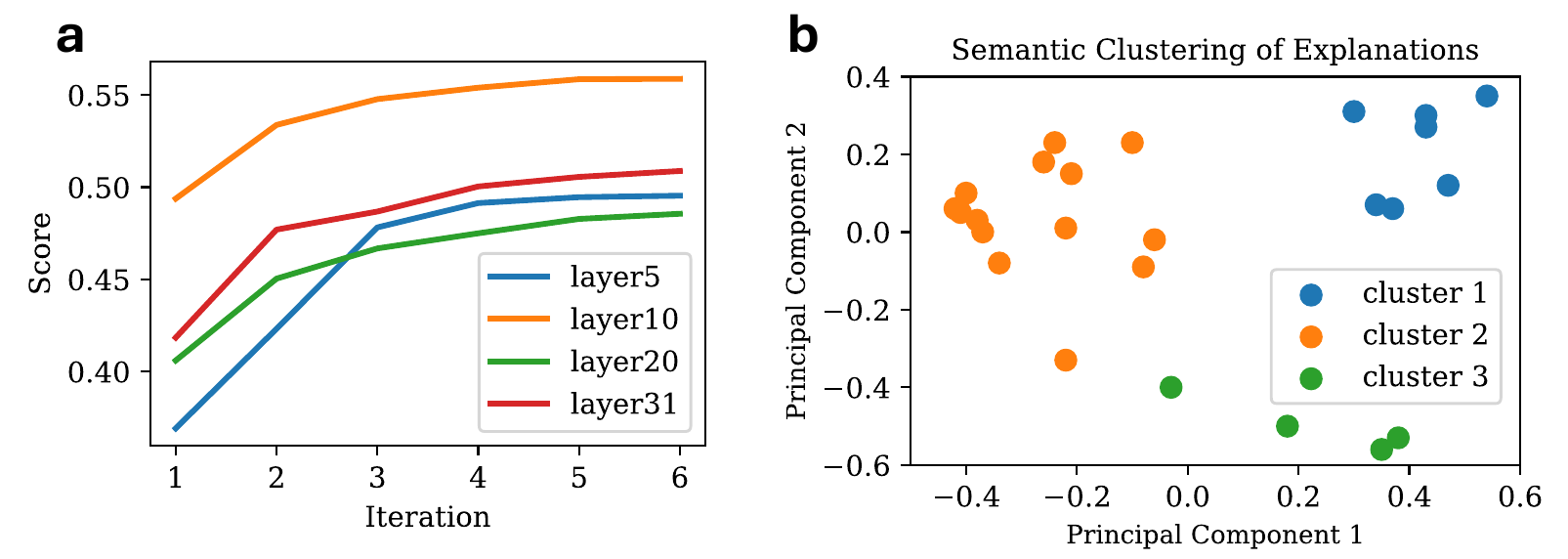}
    \vspace*{-5pt}
    \caption{\textbf{a.} Iterative Score Convergence Across Neurons and Layers. \textbf{b.} Clustering forms semantic groups.}
    \label{fig:convergence}
\end{figure*}

\section{Experiments}


\subsection{Experimental Setup}

\textbf{Datasets and Models.} We use LMSYS-Chat-1M~\cite{zheng2023lmsyschat1m} and Fineweb~\cite{penedo2024the} to extract text segments that strongly activate individual neurons. LMSYS-Chat-1M contributes diverse conversational contexts, while Fineweb provides large-scale, high-quality web text. Aggregating high-activation examples from both sources yields the corpus for explanation generation and semantic analysis. We study three open-source LLMs including Llama-3.1-8B~\cite{grattafiori2024llama3herdmodels}, Gemma-2-2B~\cite{gemmateam2024gemma2improvingopen}, and Qwen3-4B~\cite{qwen3technicalreport}, and analyze neuron activations from selected layers in each model. More implementation details are provided in Appendix~\ref{app:implementation}.

\noindent \textbf{Evaluation.} We evaluate neuron interpretations along two dimensions: faithfulness and semantic coverage. For faithfulness, {\it Score} is employed to quantify how accurately an explanation predicts neuron activations across inputs, detailed in Appendix~\ref{appendix:score}. For semantic coverage, we use {\it Number} measures how many distinct semantic concepts are associated with a neuron.

\noindent \textbf{Baseline.} To evaluate the effectiveness of NeuronScope, we systematically compare it against its variant without Refinement Agent. To ensure fair comparison, all other stages of the framework remain identical. This comparison demonstrates the improvement introduced by the Refinement Agent in enhancing both the quality and semantic diversity of explanations.

\subsection{Experimental Results}

The performance of neuron-level explanations from NeuronScope and its variant without the Refinement Agent are presented in Table~\ref{tab:explanation_quality}. We have included results obtained across model families and multiple layers.
The results indicate that
\emph{incorporating the Refinement Agent consistently improves the score metric across all models and layers.}
The absolute gains remain stable (approximately 0.05--0.10) despite substantial variation in baseline performance, indicating the major contribution of the Refinement Agent on explanation quality improvement.
These improvements persist from early to deep layers, suggesting that activation-guided refinement remains effective even when neuron behavior becomes increasingly abstract and polysemantic in deeper layers.
Besides, these faithfulness gains occur without altering raw semantics.
The Number metric remains unchanged across all settings, as decomposition and clustering are identical between the ablation and full framework.
This invariance shows that refinement operates strictly within fixed semantic clusters, improving explanation precision rather than collapsing or expanding semantic modes.

Together, these results reveal the effectiveness of NeuronScope on decomposing semantics and refining explanations without polluting raw semantics. They also
highlight iterative, activation-guided refinement as a potentially valuable addition to single-pass neuron interpretation pipelines, introducing a feedback signal that supports gradual refinement. We additionally include a comparison with existing neuron interpretation methods, with the results listed in Appendix~\ref{sec:comparison}. 

\subsection{Convergence Analysis of Refinement}
We analyze the convergence behavior of our iterative explanation refinement process by tracking the explanation score across multiple iterations.
Specifically, we randomly sample five neurons from each of four layers of the LLaMA model and record their scores over five iterations, where higher scores indicate better explanation quality.
For clarity, we report the mean of the scores across neurons at each iteration.
As shown in Figure~\ref{fig:convergence} (a), explanation scores increase steadily during the early iterations across all layers.
After approximately five iterations, the improvement becomes marginal and the curves largely saturate, indicating that the refinement process has reached a stable regime.
This saturation trend is consistently observed across randomly sampled neurons and layers, suggesting that additional iterations provide limited further benefit.
Based on this observation, we conclude that five refinement iterations are sufficient to achieve stable explanation quality in our setting.





\subsection{Clustering Forms Semantic Groups}

To demonstrate the necessity of clustering after semantic decomposition, we visualize the semantic segments of neuron 2941 in the 5th layer of Llama-3.1-8B. 
We project segment embeddings into a low-dimensional space using PCA, revealing semantic duplication among atomic explanations.
As shown in Figure~\ref{fig:convergence}(b), the decomposed segments form several well-separated clusters with strong intra-cluster semantic consistency.
Notably, segments within each cluster correspond to similar semantic or structural patterns, while different clusters capture distinct activation behaviors.
This structure provides intuitive evidence that individual neurons exhibit multiple coexisting semantic modes, motivating clustering as a principled step for characterizing polysemanticity.
Additional details are provided in Appendix~\ref{appendix:Clustering}.

\section{Related Work}

A key challenge for neuron-level interpretability is polysemanticity: a single neuron often responds to multiple unrelated semantic or structural patterns \cite{olah2020zoom}. 
Yet most methods still assign a single static description \cite{bills2023language,choi2024automatic}, masking mixed semantics and weakening activation prediction.
Our approach instead disentangles these triggers into atomic factors and uses multi-agent system to iteratively improve explanations with activation feedback. 
Additional discussion is provided in Appendix~\ref{app:Related Works}.

\section{Conclusions}
In this paper, we propose NeuronScope, a multi-agent framework for interpreting polysemantic neurons in LLMs.
By iteratively disentangling neuron behaviors, NeuronScope produces precise, multi-faceted explanations that outperform single-pass baselines.
Experiments across multiple open-source LLMs show that NeuronScope uncovers hidden polysemanticity of neurons and achieves higher activation correlation scores.

\clearpage

\section*{Limitations}

While NeuronScope demonstrates improved faithfulness in interpreting polysemantic neurons, it has several limitations. First, the iterative, multi-agent refinement process introduces additional computational overhead compared to single-pass interpretation methods, which may limit scalability to very large models or exhaustive neuron-level analysis. Second, NeuronScope relies on activation-based clustering to identify semantic modes, and its performance may be sensitive to the quality and diversity of the activation samples used. Finally, our evaluation is conducted on natural language text corpora and focuses on neuron behavior under general language usage; extending NeuronScope to more task-specific benchmarks or other modalities remains an open direction for future work.

\bibliography{custom}

\clearpage
\appendix
\section{Related Work}
\label{app:Related Works}

\noindent\textbf{LLM Neuron Explanation.} A growing body of work has investigated the problem of interpreting individual neurons or internal features in large language models by assigning them natural language descriptions~\cite{zhao2024explainability,bills2023language}. 
Early approaches to representation understanding primarily relied on human-driven analysis and manual inspection~\cite{zhang2018manifold,wang2022interpretability}.
More recent methods leverage large language models to automatically summarize these activation patterns, enabling scalable neuron-level interpretation across large networks~\cite{choi2024automatic}.
Similar ideas have also been extended to features learned by sparse autoencoders, where each latent dimension is treated as an interpretable feature and analyzed as a candidate unit for semantic interpretation~\cite{shu2025surveysparseautoencodersinterpreting, he2025saif}.

Despite their scalability, most existing neuron explanation methods adopt a single-pass formulation, producing a single static description per neuron or feature and implicitly assuming that neuron behavior is dominated by one semantic concept. In practice, however, many neurons are polysemantic, activating in response to multiple distinct and sometimes unrelated semantic or structural patterns~\cite{olah2020zoom}, making single descriptions prone to vague generalizations that poorly predict activation behavior.

While some approaches attempt to alleviate this issue through heuristics such as sampling or ranking candidate explanations, they do not explicitly model the multi-concept structure of individual neurons or provide a principled mechanism for disentangling mixed semantics, a challenge also highlighted in prior work on monosemantic feature representations~\cite{bricken2023monosemanticity}. Moreover, neuron explanation is typically formulated as a one-shot generation problem, lacking iterative validation against activation behavior.

In contrast, our work reframes neuron explanation as an iterative, activation-guided process.
By decomposing polysemantic neuron behavior into atomic components and refining explanations via activation feedback,
our approach enables more faithful interpretation beyond single-pass methods.

\vspace{3pt}
\noindent\textbf{Agents for Explainability.}
Recent work has increasingly explored the use of agent-based or multi-agent frameworks to support explainability in complex machine learning systems~\cite{ciatto2020agent,ciatto2020abstract}. Rather than relying on a single monolithic model to produce explanations, these approaches decompose the explanation process into multiple specialized roles, such as hypothesis generation, verification, critique, or refinement. By structuring explanation as a sequence of interacting reasoning steps, agent-based methods aim to improve the coherence, reliability, and interpretability of generated explanations.

Within this paradigm, MAIA~\cite{shaham2024multimodal} proposes a vision-language agent that coordinates a suite of interpretability tools to automate the analysis of computer vision models, iteratively synthesizing inputs and selecting representative examples to characterize behaviors such as feature selectivity and failure modes. In the context of language models, KnowThyself~\cite{prasai2025knowthyself} introduces an agent-based assistant that unifies diverse interpretability methods within a chat-oriented interface, enabling natural language queries over model behavior.

\section{Implementation Details}
\label{app:implementation}
All experiments were conducted on a single NVIDIA A100 GPU. We first collected highly activated text exemplars by recording neuron activations in response to natural language inputs, as shown in Table~\ref{tab:experimental_setup} for the selected LLMs and Transformer layers used in the analysis.
For each selected layer, we analyzed 200 neurons with the highest activation frequency across the corpus.
In our multi-agent framework, different stages were implemented using different large language models. 
The Hypothesis Agent and Decomposition Agent both used GPT-4o~\cite{openai2024gpt4technicalreport}. 
The Clustering Agent employed text-embedding-3-small~\cite{openai2024gpt4technicalreport} to compute explanation embeddings and applied HDBSCAN~\cite{McInnes2017} with a minimum cluster size of 2. For each cluster, we selected the explanation closest to the cluster centroid as its representative. 
The Refinement Agent used GPT-4o with up to 5 refinement iterations, generating 8 candidate explanations per iteration and selecting the highest-scoring one. 
Candidate neuron descriptions were scored using a fine‑tuned simulator model based on Llama‑3.1‑8B‑Instruct, which predicts neuron activations given a description and input on a held-out set of text exemplars.

\begin{table}[t]
\centering
\caption{This table outlines the experimental setup, detailing the selected open-source LLMs and the Transformer layers used for multi-semantic analysis.}
\label{tab:experimental_setup}
\begin{tabular}{c c}
\toprule
LLMs  & Layers \\
\midrule
Llama-3.1-8b & 5, 10, 20, 31 \\
Gemma-2-2b   & 5, 9, 18, 25  \\
Qwen3-4b     & 5, 9, 18, 25  \\
\bottomrule
\end{tabular}
\end{table}

\section{Formal Definition of the Score Metric}
\label{appendix:score}

We quantify the faithfulness of a candidate explanation $E$ for a neuron $n$ using an activation-prediction paradigm.
Given a held-out dataset $\mathcal{D}$, we employ a simulator that predicts the activation of neuron $n$ conditioned on the explanation $E$.
In practice, $\mathcal{D}$ corresponds to a validation split such as $\mathcal{D}_{\mathrm{val}}$.

Formally, for each input $x \in \mathcal{D}$, the simulator produces a predicted activation $\hat{a}_n(x \mid E)$, while the true activation is given by $a_n(x)$. 
We aggregate predicted and true activations over all inputs to obtain two aligned sequences:
\begin{align}
\hat{\mathbf{a}}_n(E; \mathcal{D}) &= \{\hat{a}_n(x \mid E)\}_{x \in \mathcal{D}} \\
\mathbf{a}_n(\mathcal{D}) &= \{a_n(x)\}_{x \in \mathcal{D}}
\end{align}

The Score of explanation $E$ is then defined as the Pearson correlation between the predicted and true activations:
\begin{align}
\mathrm{Score}(E, \mathcal{D}) = \mathrm{Corr}\bigl(\hat{\mathbf{a}}_n(E; \mathcal{D}),\, \mathbf{a}_n(\mathcal{D})\bigr)
\end{align}
where $\mathrm{Corr}(\cdot,\cdot)$ denotes the Pearson correlation coefficient. 
A higher score indicates stronger alignment between the activation behavior implied by the explanation and the actual neuron behavior on unseen data.

\section{Separation Improves Semantic Purity}
\label{appendix:Purity}

To empirically examine whether semantic entanglement exists in original neuron explanations, we design the following experimental procedure. This case study is conducted on neuron 6025 in the 5th layer of the Llama model. After the separation and clustering process, three representative base explanations are obtained, and their textual descriptions are used as semantic references. These base explanations characterize the neuron’s activation patterns from different semantic perspectives, and their original descriptions are as follows:

\begin{enumerate}
\item
\begin{itemize}[leftmargin=1pt, topsep=-3pt, itemsep=1pt, parsep=1pt]
This neuron activates when tokens contain specific fields or attributes being accessed.
\end{itemize}

\item
\begin{itemize}[leftmargin=1pt, topsep=-3pt, itemsep=1pt, parsep=1pt]
This neuron activates when it encounters specific programming or scripting keywords such as "Select", "fields", "(/", and "part".
\end{itemize}

\item
\begin{itemize}[leftmargin=1pt, topsep=-3pt, itemsep=1pt, parsep=1pt]
This neuron activates when tokens appear in contexts involving function names and their parameters in code snippets.
\end{itemize}
\end{enumerate}

Subsequently, we encode the above base explanation texts using a Sentence Transformer model (specifically, all-MiniLM-L6-v2) to obtain the corresponding semantic reference embeddings. Meanwhile, we segment the original neuron explanations at the sentence level and map each sentence into the same embedding space using the same encoding procedure. By computing the cosine similarity between each sentence embedding and the three semantic reference embeddings, we assign each sentence to the semantic reference with the highest similarity and visualize the results using different colors. Specifically, sentences aligned with semantic reference 1 are marked in orange, those aligned with semantic reference 2 in blue, and those aligned with semantic reference 3 in green.

\begin{figure*}[t]
    \centering
    \includegraphics[width=0.8\linewidth]{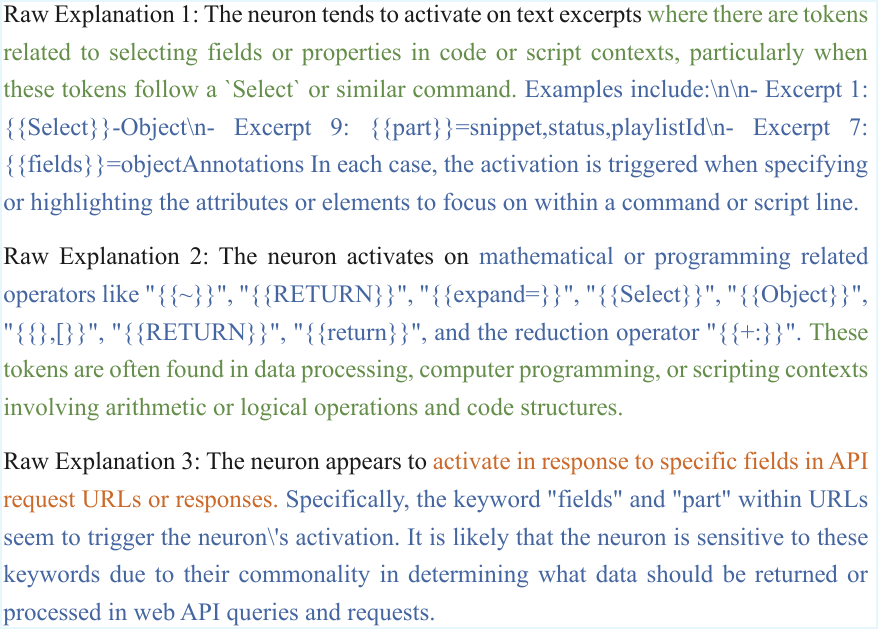}
    \caption{Semantic mixing in original neuron explanations. Different colors highlights distinct semantic categories, illustrating semantic impurity before separation.}
    \label{fig:semantic_separation}
\end{figure*}

As shown in Figure~\ref{fig:semantic_separation}, different sentences within the same original explanation exhibit clear differences in their alignment with the semantic references. This phenomenon, where a single original explanation simultaneously aligns with multiple semantic references, provides intuitive evidence that original neuron explanations are not semantically homogeneous but instead exhibit significant semantic entanglement, thereby motivating the necessity of the subsequent semantic separation operation.

\section{Illustrating Polysemantic Structure via Clustering}
\label{appendix:Clustering}

To validate the necessity of performing clustering after semantic decomposition, we select neuron 2941 in the 5th layer of the Llama model as a representative case and analyze its clustering results. Specifically, we embed the semantic segments associated with this neuron and examine their distribution in the embedding space. We observe that segments sharing similar semantic and structural characteristics exhibit clear aggregation patterns, while different types of semantic segments remain well separated. This observation indicates that the resulting clusters reveal the polysemantic nature inherent to the neuron, thereby providing intuitive evidence for the necessity of conducting polysemantic analysis at the neuron level.

In the experiment, we first collect the semantic segments obtained after decomposing the target neuron. To obtain a unified semantic representation, we use the OpenAI Embedding API to map each semantic segment into a high dimensional vector representation. After obtaining the embedding vectors, we directly perform clustering in the original high-dimensional embedding space using the HDBSCAN algorithm. Following the clustering process, in order to intuitively visualize the distribution of different semantic segments and their cluster structures in the representation space, we further apply principal component analysis (PCA) to reduce the dimensionality of the high dimensional embeddings and project them into a two dimensional space for visualization.

As shown in Figure~\ref{fig:convergence} (b), the two dimensional visualization of the semantic segments in the embedding space exhibits a clear clustering structure. Different clusters are well separated in the low-dimensional representation, while samples within the same cluster are highly concentrated in the vector space, indicating that the embedding representations effectively capture both the similarities and differences among semantic segments. These results demonstrate that clustering neuron activation segments in the embedding space can effectively reveal the coexisting polysemantic structure within a single neuron.

Specifically, the identified clusters correspond to three distinct activation patterns of the neuron. Cluster 1 primarily captures high-frequency words that undergo intentional and localized orthographic modifications, such as character insertions, substitutions, the use of diacritic markings, or unconventional capitalization, while still preserving a clearly identifiable base form. Cluster 2 is characterized by text segments containing highly irregular or non-standard combinations of letters or characters, which lack an obvious base form and deviate sharply from conventional language patterns. Cluster 3, in contrast, corresponds to novel and uninterrupted sequences of vowels or consonants occurring within words, with particularly strong activations observed when such sequences align with word boundaries, such as prefixes or suffixes.

\clearpage
\section{Agent Prompts}
\label{sec:prompts}

\begin{tcolorbox}[
    enhanced,
    breakable,
    colback=orange!10!white, colframe=blue!5!black,
    arc=2mm,
    boxrule=1pt,
    title={\bfseries Hypothesis Agent},
    coltitle=white,
    attach boxed title to top left={yshift=-2mm, xshift=3mm},
    boxed title style={enhanced, colback=blue!5!black, colframe=blue!5!black, arc=2mm, boxrule=0pt},
    top=0.5mm, left=1mm, right=1mm, bottom=0.5mm,
]
\setlength{\parskip}{2pt}
\small
\vskip8pt

You are a meticulous AI researcher conducting an important investigation into a specific neuron inside a language model that activates in response to text excerpts. Your overall task is to describe features of text excerpts that cause the neuron to strongly activate.

You will receive a list of text excerpts on which the neuron activates. Tokens causing activation will appear between delimiters like {{this}}. Consecutive activating tokens will also be accordingly delimited {{just like this}}. If no tokens are highlighted with {{}}, then the neuron does not activate on any tokens in the excerpt.

Note: Neurons activate on a word-by-word basis. Also, neuron activations can only depend on words before the word it activates on, so the description cannot depend on words that come after, and should only depend on words that come before the activation.

\end{tcolorbox}

\begin{tcolorbox}[
    enhanced,
    breakable,
    colback=orange!10!white, colframe=blue!5!black,
    arc=2mm,
    boxrule=1pt,
    title={\bfseries Decomposition Agent},
    coltitle=white,
    attach boxed title to top left={yshift=-2mm, xshift=3mm},
    boxed title style={enhanced, colback=blue!5!black, colframe=blue!5!black, arc=2mm, boxrule=0pt},
    top=0.5mm, left=1mm, right=1mm, bottom=0.5mm,
]
\setlength{\parskip}{2pt}
\small
\vskip8pt
You are an expert at analyzing neuron activation descriptions.

Your task is to split a single textual description into multiple
independent semantic activation conditions.

RULES:

1. Identify all distinct semantic activation conditions.

2. If the description contains multiple activation contexts, split them.

3. Each output sentence MUST be a complete, standalone sentence.

4. Every sentence MUST explicitly include the subject:
"This neuron activates when ..." or "This neuron fires when ..."

5. Do not shorten or rewrite the semantics. Only split and normalize.

6. If there is only one semantic meaning, return only one sentence.

7. Return the result as a JSON array of strings.

\end{tcolorbox}

\begin{tcolorbox}[
    enhanced,
    breakable,
    colback=orange!10!white, colframe=blue!5!black,
    arc=2mm,
    boxrule=1pt,
    title={\bfseries Refinement Agent},
    coltitle=white,
    attach boxed title to top left={yshift=-2mm, xshift=3mm},
    boxed title style={enhanced, colback=blue!5!black, colframe=blue!5!black, arc=2mm, boxrule=0pt},
    top=0.5mm, left=1mm, right=1mm, bottom=0.5mm,
]
\setlength{\parskip}{2pt}
\small
\vskip8pt

Requirement: 

Generate a new interpretation by applying meaningful refinements to the highest-scoring historical interpretation. 

Output format: 

Return only the new prompt text, no additional explanations or details.

Historical interpretations and their scores (ranked from highest to lowest):

\[
\begin{aligned}
\mathcal{H}^{(t)} &= \{ (E^{(i)}, \mathrm{Score}^{(i)}) : i = 0, \ldots, t \} 
\end{aligned}
\]

\end{tcolorbox}

\setlength{\parindent}{0pt}
\section{From Raw to Optimized Explanations}

In this appendix, we present a complete example illustrating the evolution of neuron explanations through our multi-agent pipeline. This example highlights how the raw outputs are progressively segmented, summarized, and finally optimized into a concise, interpretable explanation.

\vspace{0.5em} 
\textbf{Neuron 8893 in Layer 5 of the LLaMA Model:}
\vspace{0.5em} 

\textbf{Raw Explanations:}  
                                                        
[1] The neuron strongly activates in text segments where there is a mention or context related to calculation
 or computation. Specifically, it activates when phrases like "Calculate", "Compute", or similar terms related to performing arithmetic or numerical operations appear. This is evident as these words are often highlighted across the excerpts, suggesting the neuron's sensitivity to these computational or arithmetic-related contexts.  

[2] The neuron activates in contexts where it seems to be at the beginning of a line of comments or code, typically before a comment, in-line comment, or leading certain code segments in programming snippets. The activations suggest it often highlights opening delimiters for blocks, new code sections, or variable or function 
declarations, specifically when they occur at the start of a line. 

[3] The neuron seems to activate in response to incomplete or faulty code snippets provided in programming contexts, especially where comments or placeholders might be involved or where there is an abrupt transition or incorrect logic, as seen with the use of symbols like \texttt{\#} or empty/oddly placed \texttt{\{\{ \}\}}. It appears to highlight areas where the functionality intended in the code segment may not have been properly implemented or is merely indicated by a placeholder or comment that implies something should follow but doesn't, potentially identifying a spot that needs attention or completion.

\vspace{0.5em} 

\textbf{Split Explanations:}  

[1] This neuron activates when there is a mention or context related to calculation or computation in a text 
segment.

[2] This neuron activates when phrases like "Calculate", "Compute", or similar terms related to performing ar
ithmetic or numerical operations appear.

[3] This neuron activates when it encounters tokens related to sequence found in programming contexts.  

[4] This neuron activates when it encounters the symbol '\#'.

[5] This neuron activates when it encounters the closing curly brace '\}'.  

[6] This neuron activates when it encounters the concatenation symbol + when used with syntax specific to coding such as \{\{ \}\} or \{\{ + \}\}.

[7] This neuron activates when it recognizes specific coding constructs or syntax that typically appear in pr
ogramming tasks or examples.

[8] This neuron activates when the '\#' symbol is placed in comments within blocks of code. 

[9] This neuron activates when there is a comment starting with '\#' in the middle of a code structure after previously defining or loading data, often seen in examples where the code proceeds further functionality, such as defining, creating, or iterating using imported modules or self-defined functions.

[10] This neuron activates when it is at the beginning of a line of comments in programming snippets.  

[11] This neuron activates when it is at the beginning of a line of code in programming snippets.            

[12] This neuron activates when it appears before a comment in programming snippets.                         
[13] This neuron activates when it appears before an in-line comment in programming snippets.      

[14] This neuron activates when it is leading certain code segments in programming snippets. 

[15] This neuron activates when opening delimiters for blocks occur at the start of a line in programming snippets.

[16] This neuron activates when new code sections occur at the start of a line in programming snippets.

\vspace{0.5em} 

\textbf{Representative Explanations:}  

[1] This neuron activates when the code uses symbols like \# or contains empty or oddly placed \{\{ \}\} in programming contexts.

[2] This neuron activates when encountering the plus symbol '+', which is commonly used in version control sy
stems like Git to denote additions.

[3] This neuron activates when the '\#' symbol appears in Python code comments.

[4] This neuron activates when it encounters faulty code snippets in programming contexts.

[5] This neuron activates when it encounters incomplete code snippets in programming contexts.

[6] This neuron activates when it is at the beginning of a line of comments in programming snippets.

\vspace{0.5em} 

\textbf{Final Explanation:}  

[1] This neuron activates in programming contexts where code includes isolated symbols such as \#, or features unusual positioning or absence of content within \{\{ \}\} delimiters.

[2] This neuron achieves its highest activation when detecting the plus symbol '+', particularly in code diffs from version control systems like Git, where '+' precisely indicates newly added lines or content.

[3] This neuron activates in response to the '\#' character specifically marking the start of a comment within Python code.

[4] This neuron exhibits heightened activation when encountering programming code with syntactic or logical flaws, particularly within contexts that emphasize error identification or code correction.

[5] This neuron is maximally activated by standalone, unfinished programming code fragments that lack complete syntax or contextual closure.

[6] This neuron activates precisely at the initial character of a new line that marks the commencement of a comment in programming code, distinguishing it from preceding code or whitespace.

\vspace{0.5em} 
This example demonstrates the pipeline's ability to transform noisy and overlapping raw outputs into a clear, human-interpretable explanation that captures the neuron's key semantic behavior.


\setlength{\parindent}{0pt}
\section{Comparison with Other Neuron Explanation Methods}
\label{sec:comparison}

We compare NeuronScope with one existing neuron explanation method on the Llama-3.1-8B model.
The baseline neuron descriptions are generated following the approach introduced in \textit{Scaling Automatic Neuron Description}.The detailed results are summarized in Table~\ref{table:case-studies}.

\begin{table*}[!htp]
    \setlength{\belowcaptionskip}{0pt}
    \setlength{\tabcolsep}{3pt}  
    \footnotesize
    \centering
    \vspace{-5pt}
    \caption{
    \colorbox{secondbg}{\textbf{Blue}}: first semantics,
    \colorbox{bestbg}{\textbf{Red}}: second semantics,
    \colorbox{thirdbg}{\textbf{Green}}: third semantics.}
    \scalebox{0.95}{
    \begin{tabular}{
        p{2cm}   
        >{\raggedright\arraybackslash}m{7cm}  
        >{\raggedright\arraybackslash}m{8cm}  
    }
        \toprule
        \textbf{Layer / Neuron} & \textbf{Description by baseline} & \textbf{Description by NeuronScope (Ours)} \\
        \toprule

        5 / 10873 &
        The neuron appears to activate on words that are closely associated with brand names, trademarks, or unique identifiers. Specifically, the pattern suggests that the neuron activates on: Brand names or company names embedded within the text (e.g., RealE{{state}}, New{{York}}, american{{standard}}, home{{depot}}). It seems to particularly focus on parts of these names likely due to their role in recognizing and differentiating brands or labeled entities within a larger text. The activation looks to be sensitive to context preceding these tokens since it activates regardless of how the name is used (e.g., within URLs or as part of a longer domain name). &
        \colorbox{secondbg}{\parbox{\linewidth}{This neuron activates when a token serves as a unique core segment of a compound word in digital branding or domain names, specifically to maximize distinctiveness, memorability, and brand identity.}}
        \colorbox{bestbg}{\parbox{\linewidth}{This neuron robustly activates upon detecting highly recognizable compound proper nouns or brand names formed by seamlessly blending two or more distinct words into a unified lexical entity.}}
        \colorbox{thirdbg}{\parbox{\linewidth}{This neuron activates in response to words or phrases conventionally expressed in camel case, PascalCase, or similar concatenated syntactic forms standard in programming and technical nomenclature.}}
        \\ \midrule

        5 / 12085 &
        The neuron appears to activate on tokens that are part of the names of specific U.S. cities, particularly when the names occur in contexts that suggest geographic locations or institutions associated with these cities. The cities identified in the excerpts include Corvallis, Fort Collins, Auburn, Rolla, Cookeville, Pullman, Stillwater, and Orono. In these cases, the city name is often preceded by contextual clues like a university, a state or abbreviation (e.g., OR, MO, TN), or an institution related to the location. &
        \colorbox{secondbg}{\parbox{\linewidth}{This neuron is most strongly activated by the appearance of the token "Rolla" in any part of the text, consistently producing significant activation patterns across diverse contexts.}}
        \colorbox{bestbg}{\parbox{\linewidth}{This neuron activates in response to mentions of city or town names directly linked to notable institutions, events, or characteristics unique to those locations.}}
        \colorbox{thirdbg}{\parbox{\linewidth}{This neuron activates in response to tokens corresponding to names of any towns, cities, or populated places worldwide, irrespective of context, language, or script.}}
        \\ \midrule

        10 / 6064 &
        The neuron strongly activates on specific instances of the words "common" (including variations like "commonality"), "traits", "types", and "thread" when these words are used in the context of identifying shared characteristics, patterns, or threads among a group of people, events, or things. The activations are associated with discussions or descriptions about shared or typical attributes that might be found when analyzing or categorizing subjects, often as part of a comparative, analytical, or descriptive narrative. &
        \colorbox{secondbg}{\parbox{\linewidth}{This neuron activates when the word 'traits' highlights defining characteristics, patterns, or unifying features that clearly demonstrate fundamental similarities among members of a group, such as people, events, or objects.}}
        \colorbox{bestbg}{\parbox{\linewidth}{This neuron activates when 'common' denotes attributes, features, or traits that are clearly and simultaneously possessed or recognized by multiple distinct entities, groups, or events.}}
        \colorbox{thirdbg}{\parbox{\linewidth}{This neuron activates in response to explicit references to shared features, common traits, general types, recurring characteristics, or an overarching unifying thread that connects or associates multiple items or concepts.}}
        \\ \midrule
        
        20 / 7122 &
        The neuron activates on common auxiliary verbs "is" and "be," as well as the pronoun "it." These activations occur in grammatical constructions involving declarative or imperative statements, explanations, descriptions, instructions, or potentially passive voice and idiomatic expressions. The activating excerpts suggest that the neuron's activation is tied to the contextual functionality of these words, especially when they are used to express or clarify states of being, necessity, future intent, or when situated in broader narrative or explanatory contexts. &
        \colorbox{secondbg}{\parbox{\linewidth}{This neuron activates specifically in response to verbs that assert, clarify, or strongly emphasize necessity or obligation within a statement, especially when these verbs are central to conveying the statement's core meaning or intent.}}
        \colorbox{bestbg}{\parbox{\linewidth}{This neuron strongly activates when 'is', 'be', or 'it' appear in contexts where they are used to explicitly describe, define, or characterize something.}}
        \colorbox{thirdbg}{\parbox{\linewidth}{This neuron activates when the word "be" is used as the main verb in a declarative sentence or formal assertion.}}
        \\ \midrule

        10 / 12241 &
        The neuron activates on words that are part of a sequence or pairing with "series", "combo", "succession", "and", or other conjunctions implying combinations or groups. The context typically involves descriptions or instructions where items are combined, performed in a sequence, or linked together in a series for enhanced effect or clear understanding. &
        \colorbox{secondbg}{\parbox{\linewidth}{This neuron activates when the context presents instructions or descriptions that necessitate executing actions in a precisely specified, clearly articulated sequence.}}
        \colorbox{bestbg}{\parbox{\linewidth}{This neuron activates when "combo" refers specifically to the intentional, strategic pairing of rabadon and void staff to achieve maximal, synergistic amplification of power or effectiveness in a scenario.}}
        \colorbox{thirdbg}{\parbox{\linewidth}{This neuron activates when the text contains "{{succession}}", referring to a sequence or series of actions.}}
        \\ 

        \bottomrule
    \end{tabular}}
    \label{table:case-studies}
    \vspace{-10pt}
\end{table*}

\end{document}